\documentclass[letterpaper]{article} 
\usepackage{aaai2026}  
\usepackage{times}  
\usepackage{helvet}  
\usepackage{courier}  
\usepackage[hyphens]{url}  
\usepackage{graphicx} 
\usepackage{natbib}  
\usepackage{caption} 
\usepackage{algorithm}

\usepackage{comment}
\usepackage{color}
        
\usepackage[table]{xcolor}
\usepackage{booktabs}
\usepackage{enumitem}
\usepackage{amsmath}
\usepackage{mathtools}
\usepackage{url}
\usepackage{multirow, array,makecell,xspace,bm}
\usepackage{pifont}
\usepackage{amsfonts}
\usepackage{arydshln}
\usepackage{dsfont}
\usepackage{amssymb}

\usepackage{stackengine}

\usepackage{scalerel}
\usepackage{esdiff}

\usepackage{algpseudocode}
\algnewcommand{\LineComment}[1]{\State \# #1} 

\newcommand{\cmark}{\ding{51}}%
\usepackage{newfloat}
\usepackage{listings}
\DeclareCaptionStyle{ruled}{labelfont=normalfont,labelsep=colon,strut=off} 
\floatstyle{ruled}
\newfloat{listing}{tb}{lst}{}
\floatname{listing}{Listing}
\title{See, Rank, and Filter: Important Word-Aware Clip Filtering via \\Scene Understanding for Moment Retrieval and Highlight Detection}
\author{
    YuEun Lee,
    Jung Uk Kim\thanks{Corresponding author.}
}
\affiliations{
    Kyung Hee University, Yong-in, South Korea\\
    \{dbdms8435, ju.kim\}@khu.ac.kr
}

\def\maketitlesupplementary
   {
    \newpage
       \twocolumn[
        \centering
        \Large
        \textbf{See, Rank, and Filter: Important Word-Aware Clip Filtering via \\Scene Understanding for Moment Retrieval and Highlight Detection\\
        -- \textit{Supplementary Material} --}\\
        \vspace{1.5em}
       ] 
   }

\begin{document}

\maketitle

\begin{abstract}
Video moment retrieval (MR) and highlight detection (HD) with natural language queries aim to localize relevant moments and key highlights in a video clips. However, existing methods overlook the importance of individual words, treating the entire text query and video clips as a black-box, which hinders contextual understanding. In this paper, we propose a novel approach that enables fine-grained clip filtering by identifying and prioritizing important words in the query. Our method integrates image-text scene understanding through Multimodal Large Language Models (MLLMs) and enhances the semantic understanding of video clips. We introduce a feature enhancement module (FEM) to capture important words from the query and a ranking-based filtering module (RFM) to iteratively refine video clips based on their relevance to these important words. Extensive experiments demonstrate that our approach significantly outperforms existing state-of-the-art methods, achieving superior performance in both MR and HD tasks. Our code is available at: https://github.com/VisualAIKHU/SRF.
\end{abstract}

\section{Introduction}
\label{sec:intro}

The expansion of digital devices and internet platforms has sparked growing interest in video content, resulting in exponential growth in both its volume and diversity \cite{apostolidis2021video,foo2023system}. While this vast amount of content contains valuable information, reviewing it to extract relevant parts is time-consuming \cite{apostolidis2021video}. To address this, two key tasks have emerged for finding specific clips of interest based on text queries. One is moment retrieval (MR), which aims to locate relevant moments within videos \cite{charades2017}, and the other is highlight detection (HD), which tries to identify the most important clips of the videos \cite{youtubehl2014}.

Given the similarity between MR and HD tasks in identifying important video clips, the introduction of Moment-DETR and the QVhighlights dataset \cite{momentdetr2021} has encouraged joint approaches. Moment-DETR \cite{momentdetr2021} first applied the DETR framework \cite{detr2020} for this purpose. UMT \cite{umt2022} devised a framework by incorporating audio modality, and UVCOM \cite{uvcom2024} proposed an integration module for the inter- and intra-modality interaction. TR-DETR \cite{trdetr2024} focused on task interaction during training, while TaskWeave \cite{taskweave2024} proposed a task-oriented framework for more effective MR and HD. Keyword-DETR \cite{keyworddetr2025} enhanced alignment by identifying key words in the query.

\begin{figure}[t]
    \centering
    \includegraphics[width=0.49\textwidth]{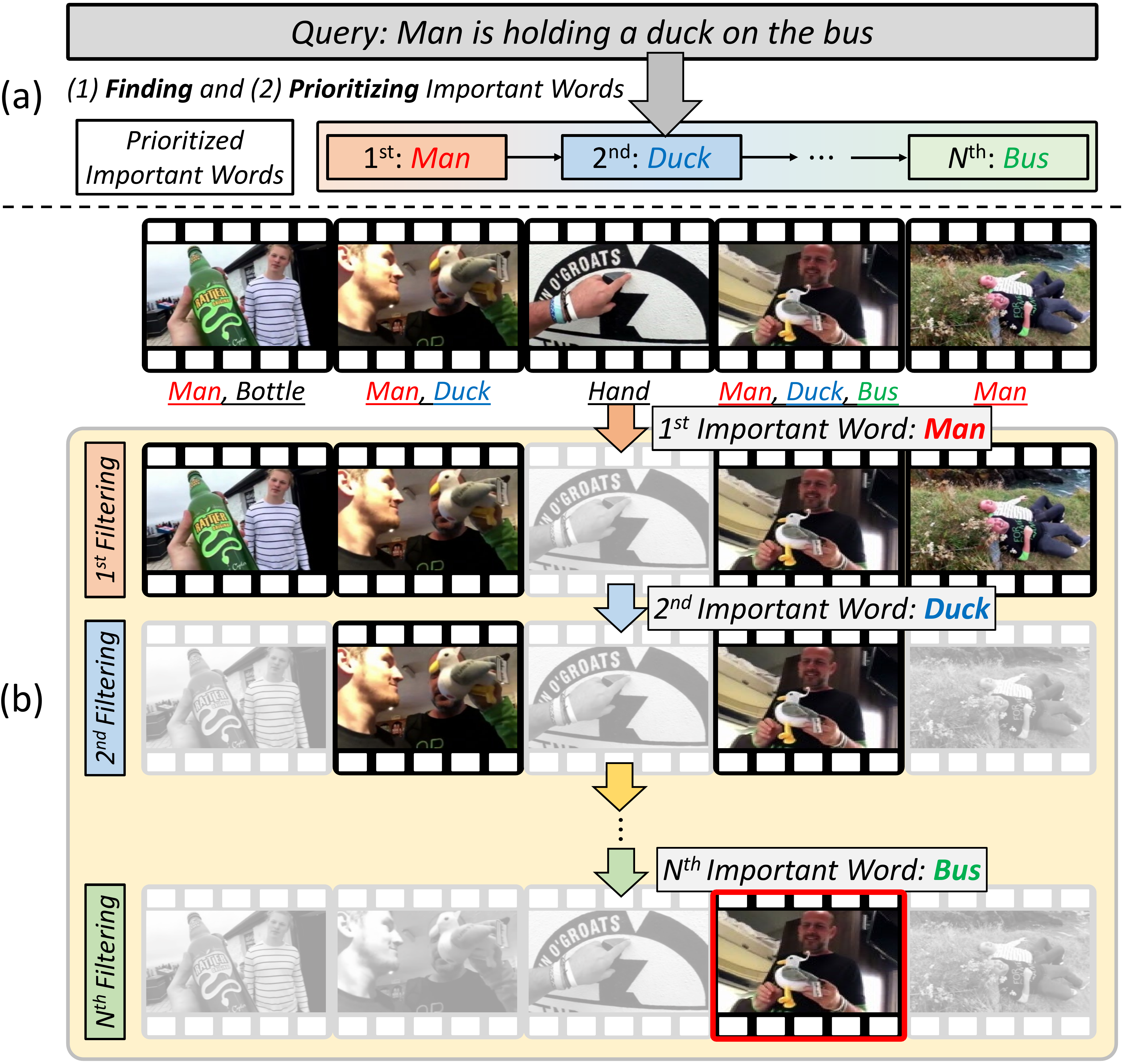}
    \caption{Conceptual illustration of our method. First, we aim to (a) find and prioritize important words in a text query, and (b) filter video clips based on the priority of the words.}
    \label{fig:1}
    \vspace{-5.0mm}
\end{figure}

While recent studies have achieved significant improvements in MR and HD considering task similarities and differences, their performance is still limited due to underutilization of the unique characteristics of text queries and video clips. To address this limitation, we draw inspiration from how humans utilize both modalities for MR and HD tasks. First, for the aspects of the text query, according to \cite{just1980theory}, given a query like `\textit{Man is holding a duck on the bus}', we naturally pre-identify important words (\textit{e.g.,} `\textit{man},' `\textit{duck},' `\textit{bus},' `\textit{hold}') and prioritize these important words before analyzing the video (Figure \ref{fig:1}(a)). 
After that, we filter the video clips related to these prioritized important words to effectively identify the clips that align with the query (Figure \ref{fig:1}(b)). However, existing methods treat the processing of the query text and video clip as a black-box, failing to prioritize the important words and thereby lacking a sufficient understanding of the query.

Second, for the aspect of the video clips, we not only interpret the visual content itself but also leverage scene understanding information \cite{vo2022contextual, buch2022revisiting}. Specifically, we analyze spatial layout, objects interactions, actions within the scene, and the temporal changes of the scene to comprehend the context while watching the video. By doing so, we gain a deeper scene understanding, which helps us find the video clip segments that are most relevant to the text query. However, existing methods rely mainly on raw visual content, limiting their ability to fully capture the scene context and properly align it with the query.

In this paper, building on these insights, we propose a novel approach that enables more fine-grained clip filtering related to the text query by fully leveraging image-text scene understanding. To this end, we address two main aspects: (\textit{i}) identifying important words in the text query and understanding video clips effectively, and (\textit{ii}) filtering video clips that are most relevant to these important words.

First, to address the issue (\textit{i}), we introduce a feature enhancement module (FEM) to identify and prioritize the important words given a text query. We also leverage Multimodal Large Language Models (MLLMs) to obtain detailed scene understanding through their rich external knowledge. By integrating MLLMs, ours further enhances the ability to interpret deeper and more complex scene understanding by combining image-text information from video clips.

Second, to address the issue (\textit{ii}), we propose a ranking-based filtering module (RFM) that refines video clips, based on the relevance of the prioritized important word. At this time, important query words are matched with the image-text information in the video clips in an iterative manner, gradually minimizing the effect of the irrelevant clips while highlighting the relevant ones. As a result, more accurate MR and HD are possible. We claim that our use of MLLMs can offer valuable insights for future MR and HD tasks by effectively integrating image-text multimodal information.

The major contributions of our paper are summarized as:

\begin{itemize} 
    \item We propose a feature enhancement module (FEM) that identifies and prioritizes important words in text queries, while enhancing detailed scene understanding through the utilization of MLLMs.
    \item  We introduce a ranking-based filtering module (RFM) that iteratively refines video clips by filtering clips based on the relevance of prioritized important words, improving moment retrieval and highlight detection.
\end{itemize}

\section{Related work}
\label{sec:related}
\subsection{Moment Retrieval and Highlight Detection}
Moment retrieval (MR) aims to find relevant video moments given a natural language query \cite{charades2017}.
There are two main approaches: proposal-based and proposal-free.
Proposal-based methods \cite{charades2017,hendricks2018localizing,sun2022you} generate candidate segments and rank them based on their match scores with the query.
In contrast, proposal-free methods \cite{li2021proposal,mun2020local,rodriguez2020proposal} directly regress start and end timestamps via video-text interaction.

\begin{figure*}[t]
    \centering
    \includegraphics[width=0.95\textwidth]{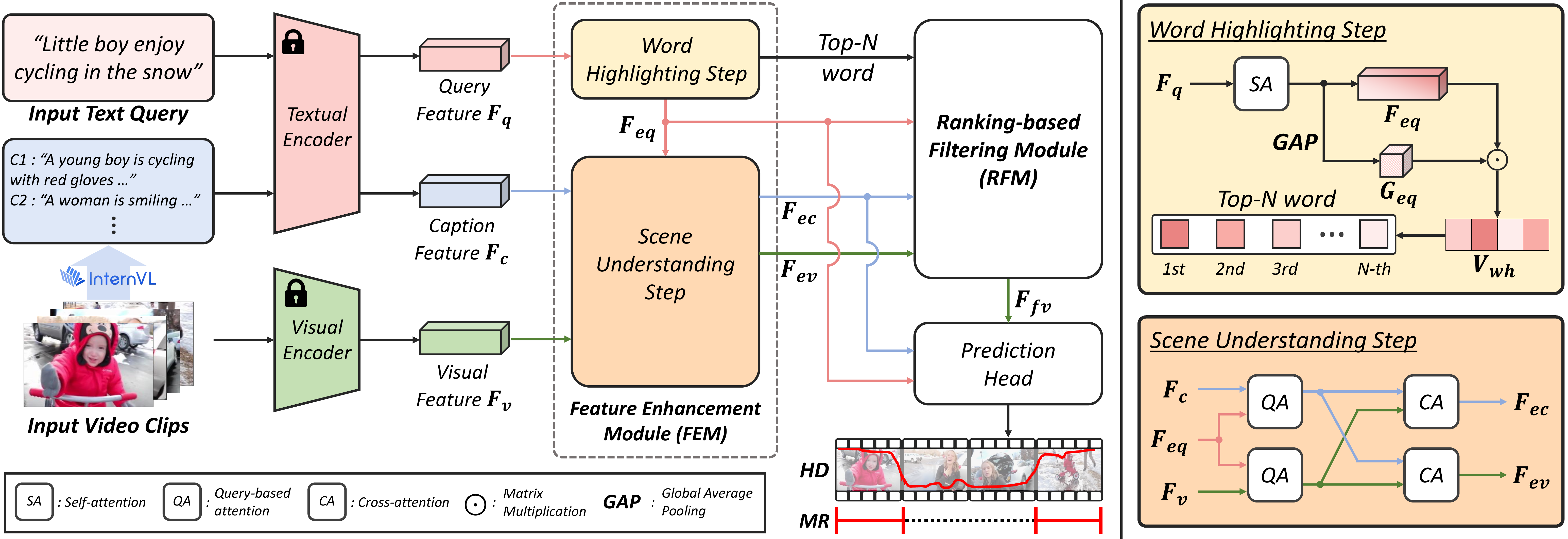}
    \caption{Overall architecture. Query, visual and caption features are prioritize important words and deepen scene understanding via feature enhancement module, then repeatedly filter irrelevant information to the query via ranking-based filtering module.
    }
    \label{fig:overall} 
    \vspace{-3.0mm}
\end{figure*}

In addition, the highlights detection (HD) aims to identify the most important video moments, \textit{i.e.,} highlights.
Early methods gave high importance scores to important moments regardless of text queries \cite{youtubehl2014,wei2022learning,badamdorj2022contrastive}.
As user preferences guide content consumption, recent HD approaches incorporate text queries to better personalize highlight selection.

Recently, MR and HD have been studied jointly.
Moment-DETR \cite{momentdetr2021} presents QVHighlights dataset and applieds DETR to both tasks.
UMT \cite{umt2022} incorporates audio alongside visual and textual inputs to enhance query understanding.
QD-DETR \cite{qddetr2023} leverages text by modeling negative video-text pairs, while TR-DETR \cite{trdetr2024} and UVCOM \cite{uvcom2024} emphasize the synergy between MR and HD. TaskWeave \cite{taskweave2024} adopts a task-centric top-down approach, and Keyword-DETR \cite{keyworddetr2025} introduces keyword-aware attention for adaptive focus.
However, existing methods fail to capture the overall context of the video and understand the semantic information of each clip, which fails to align properly with the query.
To address this issue, we propose important word-aware clip filtering framework that iteratively filters out information irrelevant to the query by integrating and leveraging scene understanding knowledge to better interpret complex scene contexts.

\subsection{Multimodal Large Language Model}
Multimodal Large Language Models (MLLMs) have evolved to meet the growing need for models that can handle multiple modalities, including not only text but also image, video, and audio. For example, CLIP \cite{clip2021} aligns visual and language modalities via contrastive learning on a large set of image-text pairs.
BLIP-2 \cite{blip2023} introduces Qformer to efficiently bridge the gap between modalities, while MiniGPT-4 \cite{minigpt2023} uses a single projection layer to match visual features to textual features.
LLaVA \cite{llava2023} improves multimodal dialogue ability by tuning instructions on multimodal data generated by GPT-4 \cite{gpt2023}.
QWen-VL \cite{qwenvl2023} uses a multi-task training strategy to fine-tune on high-resolution images.
InternVL \cite{internvl2024} scales the vision-based model and gradually aligns it with LLMs.
In this paper, we use InternVL2 to obtain rich knowledge for scene understanding of video clips.

\section{Proposed Method}
\label{sec:method}

Figure \ref{fig:overall} shows the overall framework. The text query with $L_q$ words and the untrimmed video with $L_v$ clips are processed by pre-trained modality-specific encoders (textual and visual) with three-layer feed-forward network to generate query features $F_q\in\mathbb{R}^{L_q \times d}$ and visual features $F_v\in\mathbb{R}^{L_v \times d}$. We also adopt the recent state-of-the-art Multimodal Large Language Models (MLLMs), \textit{i.e.}, InternVL2 \cite{internvl2024}, to generate captions that provide detailed semantic descriptions of each clip with up to $L_c$ words. These are encoded by textual encoder to generate caption features $F_c\in\mathbb{R}^{L_v \times d}$.

Given $F_v$, $F_q$, and $F_c$, the feature enhancement module (FEM) first identifies important words from the text query in a self-supervised manner in the word highlighting step to generate enhanced query features $F_{eq}$ and generates a word-highlighted vector $V_{wh}$ to rank important words. This module also enriches scene understanding by associating video-query and caption-query pairs in the scene understanding step, yielding enhanced visual features $F_{ev}$ and enhanced caption features $F_{ec}$. Then, based on the identified important words, the ranking-based filtering module (RFM) filters clips by emphasizing those most relevant to the query while reducing the effect of unrelated ones. This process is repeated $N$ times to obtain filtered visual features $F_{fv}$. Finally, $F_{eq}$, $F_{fv}$, and $F_{ec}$ pass through a transformer encoder-decoder to perform moment retrieval (MR) and highlight detection (HD). More details are in the following subsections.


\subsection{Feature Enhancement Module}
\label{sec:feature enhancement module}
Since the text query specifies the exact moments the user is looking for, accurately recognizing the words in the text query is essential for effective MR and HD. The video offers visual cues, while the generated clip captions provide the detailed meaning of each clip. Therefore, to effectively capture the specific moments that user is searching for based on the text query, it is important to establish strong cross-modal associations to interpret complex scene contexts.

To this end, we propose a feature enhancement module (FEM) which consists of two steps: (\textit{i}) word highlighting step and (\textit{ii}) scene understanding step. First, in the word highlighting step, as text queries contain important words and contextual information that convey the meaning of the sentence, but direct label supervision is not available, we generate enhanced query features $F_{eq} \in \mathbb{R}^{L_q \times d}$ by identifying the context through self-attention mechanism. Afterwards, the word-highlighted vector $V_{wh} \in \mathbb{R}^{L_q}$, which represents the relationship between words and between sentence-word, is generated by calculating the similarity with $G_{eq} \in \mathbb{R}^{d}$ generated through global average pooling (GAP). This process can be represented as:
\begin{gather}
    \label{eq:attention}
    F_{eq} = \text{Attention}(F_q, F_q, F_q), \\
    V_{wh} = Sim(F_{eq}, G_{eq}), \\
    Sim(X, Y) = \frac{X Y^\top}{\|X\| \|Y\|},
\end{gather}
where Attention($Q,K,V$) = Softmax$\left({Q K^\top}/{\sqrt{d}}\right) V$. By doing so, $V_{wh}$ captures the importance of words by modeling relationships between words and between words and the sentence. After learning the contextual importance by utilizing self-attention, we fine-tune the relative importance of each word in the sentence by comparing it with the global meaning. That is, each element of $V_{wh}$ can be interpreted as a score reflecting how important a specific word is in the query. If the score is high, the word is considered important, otherwise it is considered less important. After that, we rank the words to find the most important $N$ words.

Second, in the scene understanding step, similarities between $F_v$ and $F_{eq}$, $F_c$ and $F_{eq}$ are calculated to generate video-query similarity scores $A_{vq}\in \mathbb{R}^{L_v\times L_q}$ and caption-query similarity scores $A_{cq}\in \mathbb{R}^{L_v\times L_q}$, computed as:
\begin{equation}
    A_{vq} = \frac{P(F_v)\,P(F_{eq})^\top}{\sqrt{d}},\,\, A_{cq} = \frac{P(F_c) \,P(F_{eq})^\top}{\sqrt{d}},
\end{equation}
where $P(\cdot)$ is a linear projection layer. Then, row-wise softmax is applied to $A_{vq}$ and $A_{cq}$ to obtain $A^r_{vq}$ and $A^r_{cq}$, capturing the correlation between each clip or caption and all words in the text query. Column-wise softmax is also applied to obtain $A^c_{vq}$, $A^c_{cq}$, representing the correlation between a specific word in the text query and all clips or all captions.

Then, the video-to-query features $F_{v2q}$, caption-to-query features $F_{c2q}$, and the query-to-video features $F_{q2v}$ and query-to-caption features $F_{q2c}$ are calculated as follows:
\begin{gather}
    F_{v2q} = A^r_{vq} F_{eq},\,\, F_{c2q} = A^r_{cq}F_{eq},\\
    F_{q2v} = A^r_{vq} A_{vq}^{c\top} F_v,\,\, F_{q2c} = A^r_{cq} A_{cq}^{c\top} F_c.
\end{gather}

Finally, to maximize interaction between the query and $F_v$/$F_c$, we compute the query-related visual features $F_{qv}$ and query-related caption features $F_{qc}$ as follows:
\begin{gather}
    \widehat{F}_v = P(F_v\parallel F_{v2q}\parallel F_v \odot F_{v2q}\parallel F_v \odot F_{q2v}),\\
    \widehat{F}_c = P(F_c \parallel F_{c2q}\parallel F_c \odot F_{c2q}\parallel F_c \odot F_{q2c}), \\
    F_{qv} = \text{ReLU}(\text{Conv1D}(\widehat{F}_v\parallel F_{eq}^\prime)), \\
    F_{qc} = \text{ReLU}(\text{Conv1D}(\widehat{F}_c\parallel F_{eq}^\prime)),
\end{gather}
where $F_{eq}^\prime$ is sentence-level enhanced query features via a weighted sum of words \cite{huang2022video}. $(\cdot||\cdot)$ indicates concatenation and $\odot$ is Hadamard Product.

Next, we apply cross-attention to $F_{qv}$ and $F_{qc}$ to obtain enhanced visual features $F_{ev} \in \mathbb{R}^{L_v \times d}$ and enhanced caption features $F_{ec} \in \mathbb{R}^{L_v \times d}$, which can be represented as:
\begin{gather}
    F_{ev} = \text{Attention}(F_{qv}, F_{qc}, F_{qc}),\\
    F_{ec} = \text{Attention}(F_{qc}, F_{qv}, F_{qv}).
\end{gather}
This effectively integrates the two features, allowing complementary visual-textual information to enhance query-based scene understanding.


\subsection{Ranking-based Filtering Module}
\label{sec:filtering-module}
We propose a ranking-based filtering module (RFM) to emphasize video clips related to important words in the text query while suppressing unrelated ones. At this time, since the important words in the text query are diverse, our goal is to repeat this process $N$ times based on the priority of the important $N$ words to find query-relevant clips.

\begin{figure}[t]
    \centering
    \includegraphics[width=0.45\textwidth]{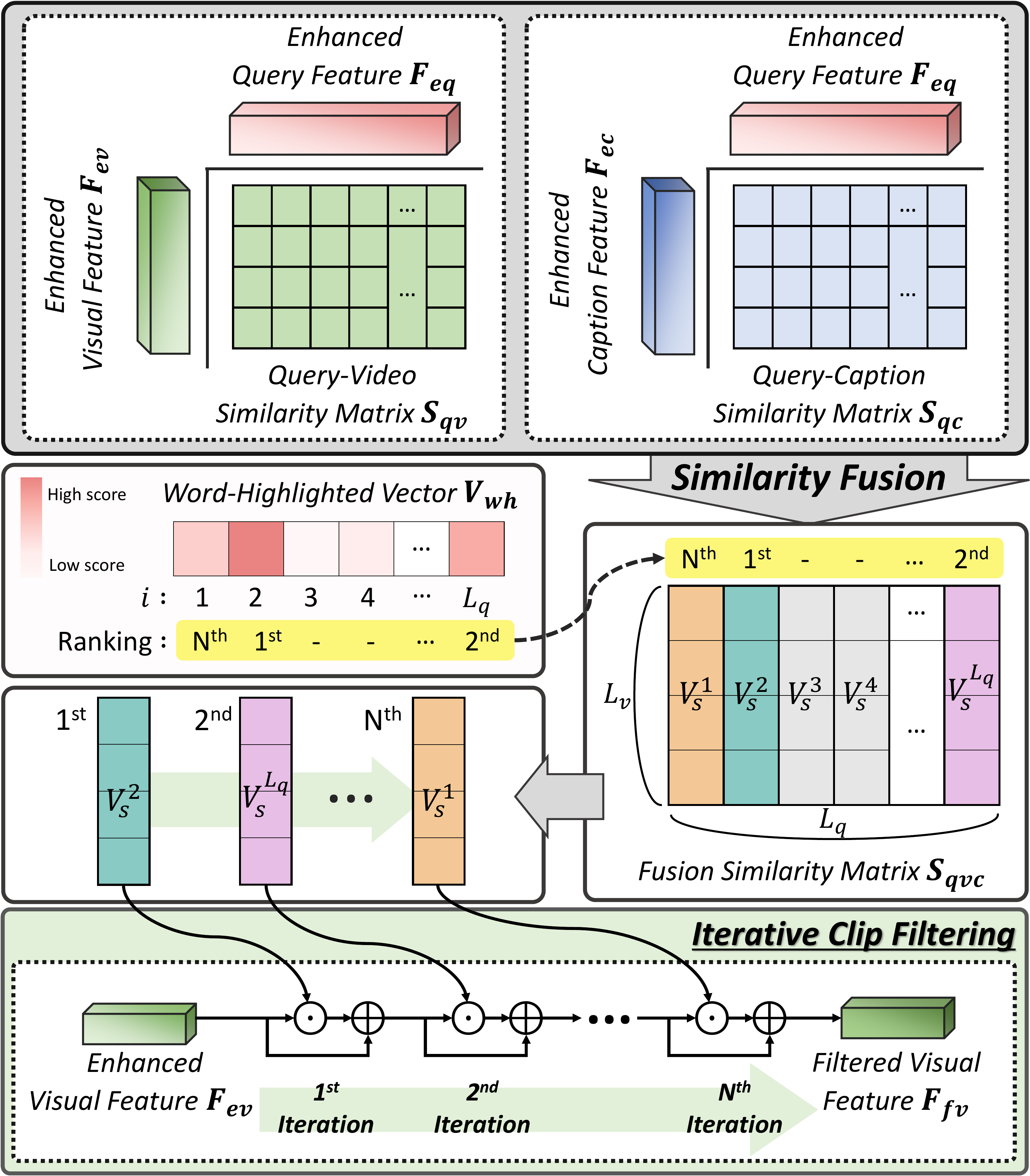}
    \caption{The detailed process of the ranking-based filtering module (RFM). Video clips are iteratively filtered based on the ranking of the most important query tokens.
    }
    \label{fig:3 RFM}
\end{figure}

As shown in Figure \ref{fig:3 RFM}, we calculate query-video and query-caption similarity matrix $S_{qv}, S_{qc}\in \mathbb{R}^{L_v \times L_q}$ to measure query relevance. Then, they are combined to obtain fusion similarity matrix $S_{qvc}$, defined as follows:
\begin{equation}
    S_{qvc} = W S_{qv} + (1-W) S_{qc},
\end{equation}
where $W$ is a learnable weight matrix that balances $S_{qv}$ and $S_{qc}$, dynamically adjusting the relative importance between video clips and captions based on the situation.

After that, the word-highlighted vector $V_{wh}$ in Eq. (4) acts as explicit prior knowledge for iterative clip filtering. The top-$N$ important words in $V_{wh}$ are used to iteratively filter the enhanced visual features $F_{ev}$. Specifically, in the first iteration ($N=1$), let $i$ be the position of the word with the highest value in $V_{wh}$. Then, we extract the $i$-th column of $S_{qvc}$ as the important word vector $V_{s}^i \in \mathbb{R}^{L_v}$, where each element indicates the similarity score between each clip and the $i$-th word in the text query. $V_{s}^i$ is applied to $F_{ev}$ with residual connection. This process is repeated $N$ times as:

\begin{equation}
    x_0 = F_{ev},\,\,\, F_{fv} =\sum_{j=1}^N x_{j-1}(1 + V_s^i).
\end{equation}

By repeating the process $N$ times, the influence of unnecessary clips is minimized, emphasizing only important clips associated with the similarity of words. The output of the iteration process is the filtered visual features $F_{fv}$. Finally, $F_{eq}$ from text query, $F_{fv}$ from the video, and $F_{ec}$ from the caption are considered to the prediction head. Note that, we follow the prediction head as \cite{trdetr2024}.

\subsection{Modal Alignment Loss}
To bridge the inherent gap between text and video arising from their different modalities, we introduce a modal alignment loss inspired by \cite{trdetr2024} that maps them to a shared semantic space. It consists of three losses: (\textit{i}) query-video alignment loss between sentence-level query features and video-level visual features, (\textit{ii}) query-clip alignment loss between sentence-level query features and clip-level visual features within a query-video pair, and (\textit{iii}) caption-clip alignment loss between sentence-level caption features and clip-level visual features within a caption-clip pair.
\begin{table*}[t]
    \renewcommand{\tabcolsep}{1.5mm}
    \centering
	\resizebox{0.999\linewidth}{!}{
		\begin{tabular}{l c cc ccccc}
            \Xhline{3\arrayrulewidth}
            \rule{0pt}{10.0pt} {\multirow{3}{*}{\qquad\qquad\qquad \bf Method}} & \multirow{3}{*}{\bf Source} & \multicolumn{5}{c}{\textbf{MR} } & \multicolumn{2}{c}{\textbf{HD} } \\ 
            \cmidrule(lr){3-4} \cmidrule(lr){5-7} \cmidrule(l){8-9}
            & & R1@0.5 & R1@0.7 & mAP@0.5 & mAP@0.75 & mAP(Avg.) & mAP & HIT@1 \\
            \midrule
            M-DETR {\scriptsize(NeurIPS'21)}   \cite{momentdetr2021} & $\mathcal{V}$ & 52.89 & 33.02 & 54.82 & 29.40 & 30.73 & 35.69 & 55.60 \\
            QD-DETR {\scriptsize(CVPR'23)} \cite{qddetr2023} & $\mathcal{V}$ & 62.40 & 44.98 & 62.52 & 39.88 & 39.86 & 38.94 & 62.40 \\
            UniVTG {\scriptsize(ICCV'23)} \cite{univtg2023} & $\mathcal{V}$ & 58.86 & 40.86 & 57.60 & 35.59 & 35.47 & 38.20 & 60.96 \\
            TR-DETR {\scriptsize(AAAI'24)} \cite{trdetr2024} & $\mathcal{V}$ & 64.66 & 48.96 & 63.98 & 43.73 & 42.62 & 39.91 & 63.42 \\
            UVCOM {\scriptsize(CVPR'24)} \cite{uvcom2024} & $\mathcal{V}$ & 63.55 & 47.47 & 63.37 & 42.67 & {43.18} & 39.74 & 64.20 \\
            Keyword-DETR {\scriptsize(AAAI'25)} \cite{keyworddetr2025} & $\mathcal{V}$ & \underline{66.86} & \underline{51.23} & \underline{67.73} & \underline{46.24} & \underline{45.69} & \underline{40.94} & \underline{64.79} \\
            \cdashline{1-9}
            \rule{0pt}{9.0pt}\cellcolor{gray!20}\bf Proposed Method & \cellcolor{gray!20}$\mathcal{V}$ & \cellcolor{gray!20}\textbf{68.09} & \cellcolor{gray!20}\textbf{52.20} & \cellcolor{gray!20}\textbf{67.81} & \cellcolor{gray!20}\textbf{46.74} & \cellcolor{gray!20}\textbf{46.54} & \cellcolor{gray!20}\textbf{42.24} & \cellcolor{gray!20}\textbf{68.22} \\
            \hline
            \rule{0pt}{9.5pt}UMT {\scriptsize(CVPR'22)} \cite{umt2022} & $\mathcal{V}+\mathcal{A}$ & 56.23 & 41.18 & 53.38 & 37.01 & 36.12 & 38.18 & 59.99 \\
            QD-DETR {\scriptsize(CVPR'23)} \cite{qddetr2023} & $\mathcal{V}+\mathcal{A}$ & 63.06 & 45.10 & 63.04 & 40.10 & 40.19 & 39.04 & 62.87 \\
            TR-DETR {\scriptsize(AAAI'24)} \cite{trdetr2024} & $\mathcal{V}+\mathcal{A}$ & 65.05 & 47.67 & {64.87} & 42.98 & 43.10 & 39.90 & 63.88 \\
            UVCOM {\scriptsize(CVPR'24)} \cite{uvcom2024} & $\mathcal{V}+\mathcal{A}$ & 63.81 & {48.70} & 64.47 & {44.01} & {43.27} & 39.79 & {64.79} \\
            Keyword-DETR {\scriptsize(AAAI'25)} \cite{keyworddetr2025} & $\mathcal{V}+\mathcal{A}$ & \underline{67.77} & \underline{50.52} & \bf{68.30} & \underline{45.88} & \underline{45.52} & \underline{41.15} & \underline{65.82} \\
            \hdashline
            \rule{0pt}{9.0pt}\cellcolor{gray!20}\textbf{Proposed Method} & \cellcolor{gray!20}$\mathcal{V}+\mathcal{A}$ & \cellcolor{gray!20}\textbf{68.87} & \cellcolor{gray!20}\textbf{52.27} & \cellcolor{gray!20}\underline{68.09} & \cellcolor{gray!20}\textbf{46.55} & \cellcolor{gray!20}\textbf{46.23} & \cellcolor{gray!20}\textbf{42.36} & \cellcolor{gray!20}\textbf{69.78} \\
            \Xhline{3\arrayrulewidth}
            \end{tabular}
        }
    \caption{Results of moment retrieval and highlight detection experiments on the QVHighlights \textit{test} set using video only ($\mathcal{V}$) and video and audio together ($\mathcal{V}+\mathcal{A}$). Best/second-best results are marked in \textbf{Bold}/\underline{underlined}.}
    \label{table:qvhighlight}
\end{table*}
First, the query-video alignment loss $\mathcal{L}_{q\text{-}v}$ is calculated as:
\begin{gather}
    \mathcal{L}_{q\text{-}v} = -\frac{1}{B} \sum_{j=1}^{B} \log \frac{\exp(Sim(G_{v_j}, G_{q_j}))}{\sum_{i=1}^{B} \exp(Sim(G_{v_i}, G_{q_j}))},
\end{gather}
where $B$ is the batch size, $G_{v_i}, G_{q_i} \in \mathbb{R}^d$ are the $i$-th global visual and query features, obtained via GAP of $F_{v_i}$ and $F_{q_i}$ respectively. $\mathcal{L}_{q\text{-}v}$ enhances global correlation of similar query-video pairs by separating them from dissimilar pairs.

Next, the query-video similarity matrix $S_{qv}$ is passed through the sigmoid and average pooling to generate $G_{qc_i}$. Then, the query-clip alignment loss $\mathcal{L}_{q\text{-}c}$ is defined as:
\begin{equation}
    \mathcal{L}_{q\text{-}c} = - \sum_{i=1}^{L_v} \left( M_i \log(G_{qc_i}) + (1 - M_i) \log(1 - G_{qc_i}) \right),
\end{equation}
where $M_i$ is a ground-truth mask, which means 1 if the $i$-th video clip is relevant to the query, and 0 otherwise. $G_{qc_i}$ is the similarity score between the global query features and the $i$-th clip-level visual features. Through $\mathcal{L}_{q\text{-}c}$, relevant and irrelevant video clips are differentiated based on the query.

Finally, the caption-clip alignment loss $\mathcal{L}_{c\text{-}c}$ is defined as:
\begin{gather}
    \mathcal{L}_{c\text{-}c} = -\frac{1}{B} \sum_{k=1}^{B} \log \frac{\sum_{j=1}^{L_v} \exp (Sim( F_{v_{kj}}, F_{{c}_{kj}}))}{\sum_{i=1}^{B} \sum_{j=1}^{L_v} \exp ( Sim( F_{{v}_{ij}}, F_{{c}_{kj}}))},
\end{gather}
where $F_{v_{kj}}, F_{c_{kj}}$ denotes the visual and caption features for the $j$-th clip of the $k$-th video.
This improves the correlation of similar caption-clip pairs in a video and better separates dissimilar pairs or similar pairs with different meanings.

Finally, the modal alignment loss $\mathcal{L}_{ma}$ is formulated as:
\begin{equation}
    \mathcal{L}_{ma} = \lambda_{q\text{-}v}\mathcal{L}_{q\text{-}v} + \lambda_{q\text{-}c}\mathcal{L}_{q\text{-}c} + \lambda_{c\text{-}c}\mathcal{L}_{c\text{-}c},
    \label{eq:modal alignment loss}
\end{equation}
where $\lambda_{q\text{-}v}$, $\lambda_{q\text{-}c}$ and $\lambda_{c\text{-}c}$ are balancing weights.
Finally, the total training loss function is formulated as follows:
\begin{equation}
    \mathcal{L}_{Total} = \mathcal{L}_{mr} + \mathcal{L}_{hd} + \mathcal{L}_{ma},
    \label{eq:total loss}
\end{equation}
where $\mathcal{L}_{mr}$ and $\mathcal{L}_{hd}$ denote the MR/HD losses from \cite{qddetr2023,trdetr2024}.

\begin{table*}[t]
    \renewcommand{\tabcolsep}{2.9mm}
    \centering
	\resizebox{0.999\linewidth}{!}{
		\begin{tabular}{l cccccccccccc}
            \Xhline{3\arrayrulewidth}
            \rule{0pt}{10.0pt}
            {\qquad\qquad\qquad \textbf{Method}} & VT & VU & GA & MS & PK & PR & FM & BK & BT & DS & Avg. \\
            \hline
            \rule{0pt}{9.5pt}LIM-S {\scriptsize(CVPR'19)} \cite{lim-s2019} & 55.9 & 42.9 & 61.2 & 54.0 & 60.3 & 47.5 & 43.2 & 66.3 & 69.1 & 62.6 & 56.3 \\
            Trailer {\scriptsize(ECCV'20)} \cite{trailer2020} & 61.3 & 54.6 & 65.7 & 60.8 & 59.1 & 70.1 & 58.2 & 64.7 & 65.6 & 68.1 & 62.8 \\
            SL-Module {\scriptsize(ICCV'21)} \cite{slmodule2021} & 86.5 & 68.7 & 74.9 & 86.2 & 79.0 & 63.2 & 58.9 & 72.6 & 78.9 & 64.0 & 73.3 \\
            UMT† {\scriptsize(CVPR'22)} \cite{umt2022} & 87.5 & 81.5 & 88.2 & 78.8 & 81.5 & 87.0 & 76.0 & 86.9 & 84.4 & 79.6 & 83.1 \\
            QD-DETR {\scriptsize(CVPR'23)} \cite{qddetr2023} & 88.2 & 87.4 & 85.6 & 85.0 & 85.8 & 86.9 & 76.4 & 91.3 & 89.2 & 73.7 & 85.0 \\
            UniVTG {\scriptsize(ICCV'23)} \cite{univtg2023} & 83.9 & 85.1 & 89.0 & 80.1 & 84.6 & 87.0 & 70.9 & 91.7 & 73.5 & 69.3 & 81.0 \\
            TR-DETR {\scriptsize(AAAI'24)} \cite{trdetr2024} & \underline{89.3} & 93.0 & 94.3 & 85.1 & 88.0 & 88.6 & 80.4 & 91.3 & 89.5 & \textbf{81.6} & 88.1 \\
            UVCOM {\scriptsize(CVPR'24)} \cite{uvcom2024} & 87.6 & 91.6 & 91.4 & \underline{86.7} & 86.9 & 86.9 & 76.9 & 92.3 & 87.4 & 75.6 & 86.3  \\
            TaskWeave {\scriptsize(CVPR'24)} \cite{taskweave2024} & 88.2 & 90.8 & 93.3 & \textbf{87.5} & 87.0 & 82.0 & \underline{80.9} & \underline{92.9} & 89.5 & \underline{81.2} & 87.3\\
            Keyword-DETR {\scriptsize(AAAI'25)} \cite{keyworddetr2025} & \bf 89.9 & \underline{93.8} & \underline{94.4} & 85.9 & \underline{89.2} & \underline{89.4} & \textbf{81.5} & 92.6 & \textbf{90.1} & 80.6 & \underline{88.7} \\
            \cdashline{1-12}
            \rule{0pt}{9.5pt}\cellcolor{gray!20}\textbf{Proposed Method} & \cellcolor{gray!20}\textbf{89.9} & \cellcolor{gray!20}\textbf{94.1} & \cellcolor{gray!20}\textbf{95.0} &
            \cellcolor{gray!20}\textbf{87.5} & \cellcolor{gray!20}\textbf{89.7} & \cellcolor{gray!20}\textbf{90.4} & \cellcolor{gray!20}{80.6} & \cellcolor{gray!20}\textbf{93.3} & \cellcolor{gray!20}\underline{89.9} & \cellcolor{gray!20}\textbf{81.6} & \cellcolor{gray!20}\textbf{89.2} \\
            \Xhline{3\arrayrulewidth}
            \end{tabular}
        }
    \caption{Results on highlight detection experiments on the TVSum. † means training with audio modality. Best/second-best results are marked in \textbf{Bold}/\underline{underlined}.}
    \label{table:tvsum}
\end{table*}

\begin{table}[t!]
    \renewcommand{\tabcolsep}{0.3mm}
    \centering
	\resizebox{1.0\linewidth}{!}{
		\begin{tabular}{l ccc}
            \Xhline{3\arrayrulewidth}
            \rule{0pt}{10.0pt}
            {\qquad\qquad\qquad\textbf{Method}} & \textbf{Feat} 
            & R1@0.5 & R1@0.7 \\
            \hline
            \rule{0pt}{9.5pt}UMT† {\scriptsize(CVPR'22)} \cite{umt2022} & VGG & 48.31 & 29.25 \\
            QD-DETR {\scriptsize(CVPR'23)} \cite{qddetr2023} & VGG & 52.77 & 31.13 \\
            TR-DETR {\scriptsize(AAAI'24)} \cite{trdetr2024} & VGG & 53.47 & 30.81 \\
            TaskWeave {\scriptsize(CVPR'24)} \cite{taskweave2024} & VGG & \underline{56.51} & \underline{33.66} \\
            Keyword-DETR {\scriptsize(AAAI'25)} \cite{keyworddetr2025} & VGG & 54.89 & 31.97 \\
            \cdashline{1-4}
            \rule{0pt}{9.5pt}\cellcolor{gray!20}\textbf{Proposed Method} & \cellcolor{gray!20}VGG & \cellcolor{gray!20}\textbf{61.51} & \cellcolor{gray!20}\textbf{37.58} \\
            \hline
            \rule{0pt}{9.5pt}QD-DETR {\scriptsize(CVPR'23)} \cite{qddetr2023} & SF+C & 57.31 & 32.55 \\
            UniVTG {\scriptsize(ICCV'23)} \cite{univtg2023} & SF+C & 58.01 & 35.65 \\
            TR-DETR {\scriptsize(AAAI'24)} \cite{trdetr2024} & SF+C & 57.61 & 33.52 \\
            UVCOM {\scriptsize(CVPR'24)} \cite{uvcom2024} & SF+C & 59.25 & 36.64 \\
            Keyword-DETR {\scriptsize(AAAI'25)} \cite{keyworddetr2025} & SF+C & \bf {61.08} & \underline{37.89} \\
            \cdashline{1-4}
            \rule{0pt}{9.5pt}\cellcolor{gray!20}\textbf{Proposed Method} & \cellcolor{gray!20}SF+C & \cellcolor{gray!20}\underline{60.97} & \cellcolor{gray!20}\bf 38.52 \\
            \Xhline{3\arrayrulewidth}
            \end{tabular}
        }
    \caption{Results on moment retrieval experiments on the Charades-STA. $\dagger$ indicates training with audio modality. Best/second-best results are marked in \textbf{Bold}/\underline{underlined}. }
    \label{table:charades}
\end{table}

\begin{table}[t]
    \renewcommand{\tabcolsep}{0.7mm}
    \centering
	\resizebox{\linewidth}{!}{
		\begin{tabular}{c cc c ccccc}
            \Xhline{3\arrayrulewidth}
            \rule{0pt}{10.0pt}\multirow{3}{*}{\bf Cap.} & \multirow{3}{*}{\bf FEM} & \multirow{3}{*}{\bf RFM} & \multicolumn{3}{c}{\textbf{MR}} & \multicolumn{2}{c}{\textbf{HD} } \\
            \cmidrule(lr){4-6} \cmidrule(l){7-8}
            & & & R1@0.5 & R1@0.7 & mAP(Avg.) & mAP & HIT@1 \\
            \midrule
            - & - & - & 66.13 & 49.74 & 43.24 & 40.11 & 64.77 \\\cdashline{1-8}
            \rule{0pt}{9.5pt}- & \cmark & - & 66.32 & 49.81 & 43.53 & 40.80 & 65.10 \\
            - & - & \cmark & 67.55 & 49.61 & 44.38 & 41.29 & 65.81 \\
            - & \cmark & \cmark & 69.42 & 51.81 & 45.56 & 41.32 & 66.06 \\
            \cmark & - & - & 69.03 & 52.71 & 46.34 & 42.23 & 67.94 \\
            \cmark & \cmark & - & 69.16 & 53.81 & 46.81 & 42.86 & 69.61 \\
            \cmark & - & \cmark & 69.42 & 53.87 & 47.43 & 42.29 & 68.00 \\\cdashline{1-8}
            \rule{0pt}{9.5pt}\cellcolor{gray!20}\cmark & \cellcolor{gray!20}\cmark & \cellcolor{gray!20}\cmark & \cellcolor{gray!20}\textbf{70.00} & \cellcolor{gray!20}\textbf{55.68} & \cellcolor{gray!20}\textbf{48.14} & \cellcolor{gray!20}\textbf{43.24} & \cellcolor{gray!20}\textbf{71.23} \\
            \Xhline{3\arrayrulewidth}
            \end{tabular}
        }
    \caption{Effect of our components (caption usage (Cap.), feature enhancement module (FEM), and ranking-based filtering module (RFM)) on the QVHighlights \textit{val} set.}
    \label{table:module_ablation}
\end{table}

\section{Experiments}
\label{sec:exp}
\subsection{Datasets and Evaluation Metrics}

\noindent\textbf{Dataset.} We use three benchmark datasets. \textbf{QVHighlights} dataset \cite{momentdetr2021} contains 10,148 content-rich YouTube videos, paired with text queries that identifies a specific highlight moment. It includes annotations for both moment retrieval (MR) and highlight detection (HD). Test annotations are hidden, and results are evaluated via the CodaLab server. \textbf{TVSum} dataset \cite{tvsum2015} includes 50 videos across 10 categories for HD. Following \cite{qddetr2023}, 80$\%$ of the dataset is used for training, and 20$\%$ is used for testing. \textbf{Charades-STA} dataset \cite{charades2017} contains 9,848 videos of indoor daily activities and 16,128 human-annotated text queries. Following \cite{qddetr2023}, 12,408 samples are used for training and 3,720 samples for testing. \\

\noindent\textbf{Evaluation Metrics.} We follow the evaluation metrics used in previous works \cite{trdetr2024,uvcom2024} for fair comparison. For QVHighlights, we measure Recall@1 (R1) at IoU thresholds of 0.5 and 0.7 for MR, and compute the average mAP (mAP@Avg) for IoU thresholds sampled at 0.05 intervals from 0.5 to 0.95. We also evaluate mAP at specific thresholds of 0.5 and 0.75 for more detailed performance comparison. For HD, we use the average precision (mAP) and HIT@1, which represents the hit rate of the highest-scoring clip. For TVSum, we evaluate HD using the top-5 mAP values. For Charades-STA, we measure R1 at IoU thresholds of 0.5 and 0.7 for MR.

\subsection{Implementation Details}
Following \cite{trdetr2024}, we extracted video, query, caption, and audio features using pre-trained models. For video, SlowFast and CLIP \cite{clip2021} were used for QVHighlights; VGG \cite{vgg2014} and SlowFast+CLIP for Charades-STA; and I3D for TVSum. Query and caption features were extracted using CLIP for QVHighlights and TVSum, and GLoVe for Charades-STA. All audio features were obtained using PANN \cite{pann2020} trained on the AudioSet \cite{audioset2017}.

All experiments were performed on an NVIDIA RTX 3090, with $\lambda_{q\text{-}v}=0.3$, $\lambda_{q\text{-}v}=0.5$, $\lambda_{c\text{-}c}=1.5$ in Eq. (\ref{eq:modal alignment loss}).
Other training settings followed TR-DETR \cite{trdetr2024}

\subsection{Experimental Results}
\noindent\textbf{Results on the QVHighlights.}
Table \ref{table:qvhighlight} shows the experimental results for MR and HD on the QVHighlights \textit{test} set. We compared state-of-the-art methods \cite{momentdetr2021, qddetr2023, univtg2023, trdetr2024, uvcom2024, keyworddetr2025, umt2022}. Our method shows superior performance across all metrics when using only video ($\mathcal{V}$). With the concatenated video and audio setting ($\mathcal{V} + \mathcal{A}$), it still outperforms other methods on most metrics. These results demonstrate the effectiveness of our method in identifying important query words and understanding video content. \\

\noindent\textbf{Results on the TVSum.} We evaluated the HD performance on the TVSum dataset. As shown in Table \ref{table:tvsum}, the overall average performance (Avg.) of our method still outperforms the existing methods in almost all categories. This highlights that our method is still effective approach for HD task. \\

\noindent\textbf{Results on the Charades-STA.} As shown in Table \ref{table:charades}, when evaluating the MR task on the Charades-STA dataset, our method outperforms state-of-the-art models in most cases, regardless of the video feature type (i.e., VGG or SlowFast+CLIP(SF+C)). These experimental results demonstrate that proposed method, which effectively understands both query text and video clips, remains effective in MR task.
\begin{figure*}[t]
    \centering
    \includegraphics[width=1.0\textwidth]{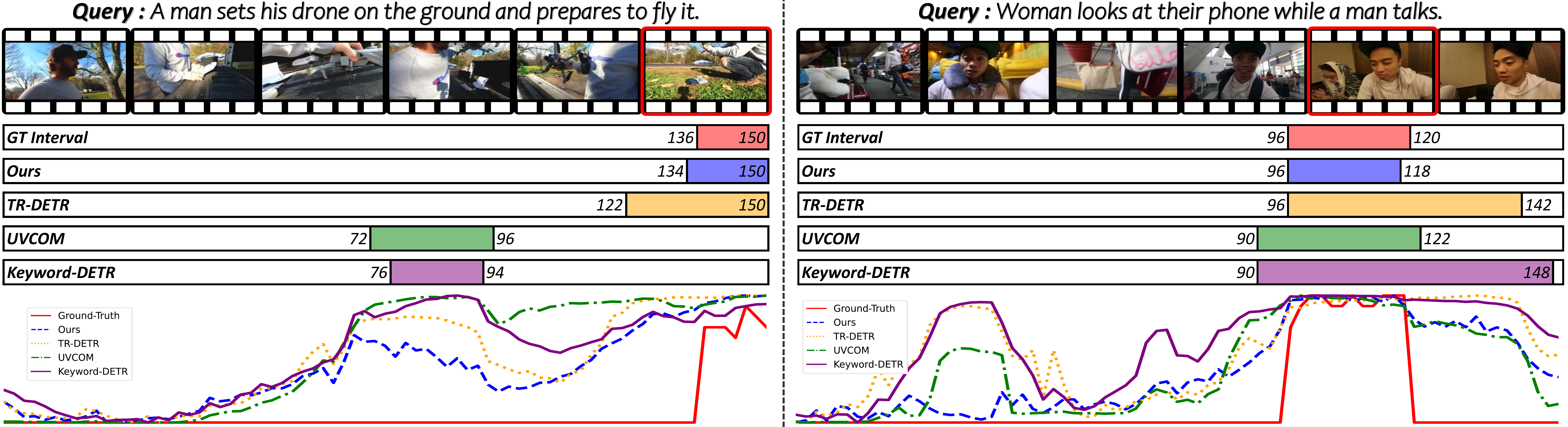}
    \caption{Visualization comparison of moment retrieval (MR) and highlight detection (HD) for the QVHighlights \textit{val} set.}
    \label{fig:visualization} 
\end{figure*}

\begin{figure}[t]
    \centering
    \includegraphics[width=0.46\textwidth]{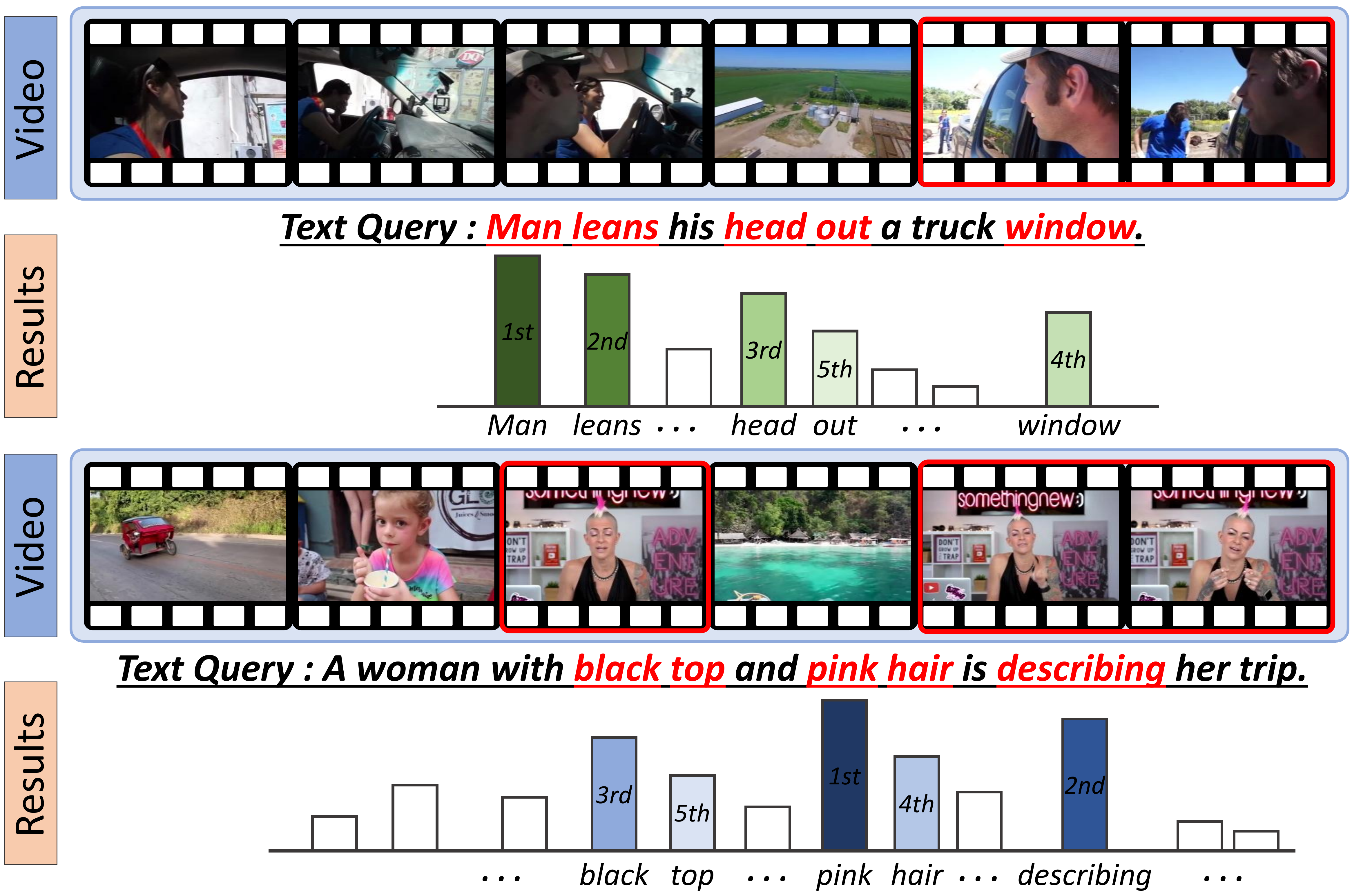}
    \caption{Visualization results of the prioritized $N$ important words on the QVHighlights \textit{val} set ($N=5$).}
    \label{fig:query-visualization}
\end{figure}

\subsection{Ablation Study}
We perform an ablation study on the QVHighlights \textit{val} set to see the effectiveness of each module in our method. As shown in Table \ref{table:module_ablation}, both FEM and RFM contribute to improved performance compared to the baseline, and their combination achieves the best results. Adding captions further enhances scene understanding and overall performance.

\subsection{Visualization Results}
\noindent\textbf{Qualitative Comparisons.} We compared our method with state-of-the-art approaches \cite{trdetr2024, uvcom2024, keyworddetr2025} on the QVHighlights \textit{val} set. As shown in Figure \ref{fig:visualization}, ours outperformed existing ones, demonstrating its superiority in achieving more accurate MR and HD predictions. \\

\noindent\textbf{Results of the Ranking Important Words.} 
Figure \ref{fig:query-visualization} shows the top-\textit{N} words from the word-highlighted vector $V_{wh}$, obtained during the word highlighting step of RFM. It demonstrates that the model prioritizes words strongly connected to other components or central to actions or objects, which effectively helps retrieve relevant clips.

\subsection{Discussions}
\label{sec:discussion}
\noindent\textbf{Effect of the Caption Information.} While leveraging captions with MLLMs is a promising recent trend, previous MR/HD studies have not yet explored integrating visual-text multimodal information to enhance video understanding. In contrast, we are the first to incorporate multimodal information in MR/HD tasks. For fair comparison, Table \ref{table:effect of caption features} reports results for adding MLLM-generated captions to SOTA models. These results demonstrate that the superiority of this study lies not in the captions, but in effectively identifying and prioritizing important query words.

\noindent\textbf{Effect of Important Word-based Iterative Filtering.} 
We evaluate the effect of the number of iterations in RFM on the QVHighlights \textit{val} set, as shown in Table~\ref{table:iter_ablation}. Increasing the number of iterations leads to performance gains by refining clip selection, with the best performance at $N=5$. At $N=7$, it slightly drops due to over-filtering but still outperforms the no-iteration baseline. \\

\begin{table}[t!]
    \renewcommand{\tabcolsep}{1.5mm}
    \centering
	\resizebox{0.99\linewidth}{!}{
		\begin{tabular}{l ccc cc}
            \Xhline{3\arrayrulewidth}
            \rule{0pt}{10.0pt}
            \multirow{3}{*}[-0.3em]{\qquad\bf Method} & \multicolumn{3}{c}{\textbf{MR} } & \multicolumn{2}{c}{\textbf{HD} } \\
            \cmidrule(lr){2-4} \cmidrule(l){5-6}
            & \makecell{R1\\@0.5} & \makecell{R1\\@0.7} & \makecell{mAP\\(Avg.)} & mAP & \makecell{HIT\\@1} \\
            \midrule
            TR-DETR {\scriptsize(AAAI'24)} & 68.90 & 51.94 & 46.60 & 42.15	& 68.13 \\
            UVCOM {\scriptsize(CVPR'24)} & 68.71 & 53.23 & 47.68 & 41.95 & 68.32 \\
            Keyword-DETR {\scriptsize(AAAI'25)} & 68.32 & 53.35 & 47.73 & 41.97 & 69.55 \\
            \cdashline{1-6}
            \rule{0pt}{9.0pt}\cellcolor{gray!20}\bf Proposed Method & \cellcolor{gray!20}\textbf{70.00} & \cellcolor{gray!20}\textbf{55.68} & \cellcolor{gray!20}\textbf{48.14} & \cellcolor{gray!20}\textbf{43.24} & \cellcolor{gray!20}\textbf{71.23} \\
            \Xhline{3\arrayrulewidth}
            \end{tabular}
        }
    \caption{Effect of the caption information on the QVHighlights \textit{val} set. InternVL2 is used for caption extraction.
}
    \label{table:effect of caption features}
\end{table}
\begin{table}[t]
    \renewcommand{\tabcolsep}{1.5mm}
    \centering
	\resizebox{0.99\linewidth}{!}{
		\begin{tabular}{c ccccc}
            \Xhline{3\arrayrulewidth}
            \rule{0pt}{10.0pt}
            \multirow{2}{*}[-0.3em]{\bf \# of Iter} & \multicolumn{3}{c}{\textbf{MR} } & \multicolumn{2}{c}{\textbf{HD} } \\
            \cmidrule(lr){2-4} \cmidrule(l){5-6}
            & R1@0.5 & R1@0.7 & mAP(Avg.) & mAP & HIT@1 \\
            \midrule
            0 & 69.16 & 53.81 & 46.81 & 42.86 & 69.61 \\\cdashline{1-6}
            \rule{0pt}{10.0pt}1 & 69.48 & 54.13 & 47.10 & 42.91 & 70.06 \\
            3 & 69.74 & 54.45 & 47.30 & 42.98 & 70.90 \\
            \cellcolor{gray!20}\textbf{5} & \cellcolor{gray!20}\textbf{70.00} & \cellcolor{gray!20}\textbf{55.68} & \cellcolor{gray!20}\textbf{48.14} & \cellcolor{gray!20}\textbf{43.24} & \cellcolor{gray!20}\textbf{71.23} \\
            7 & \bf 70.00 & 53.87 & 47.09 & 42.76 & 69.29 \\
            \Xhline{3\arrayrulewidth}
            \end{tabular}
        }
    \caption{Effect of the number of filtering iterations in the ranking-based filtering module on the QVHighlights \textit{val} set.}
    \label{table:iter_ablation}
\end{table}

\begin{table}[t!]
    \renewcommand{\tabcolsep}{1.5mm}
    \centering
	\resizebox{0.99\linewidth}{!}{
		\begin{tabular}{c ccccc}
            \Xhline{3\arrayrulewidth}
            \rule{0pt}{10.0pt}
            \multirow{2}{*}[-0.3em]{\bf MLLMs} & \multicolumn{3}{c}{\textbf{MR} } & \multicolumn{2}{c}{\textbf{HD} } \\
            \cmidrule(lr){2-4} \cmidrule(l){5-6}
            & \makecell{R1@0.5} & \makecell{R1@0.7} & \makecell{mAP(Avg.)} & mAP & \makecell{HIT@1} \\
            \midrule
            \rule{0pt}{10.0pt}
            LLaVA & \bf 70.71 & 54.90 & 47.69 & 42.80 & 69.68 \\
            \cellcolor{gray!20}\textbf{InternVL2} & \cellcolor{gray!20}70.00 & \cellcolor{gray!20}\textbf{55.68} & \cellcolor{gray!20}\textbf{48.14} & \cellcolor{gray!20}\textbf{43.24} & \cellcolor{gray!20}\textbf{71.23} \\
            \Xhline{3\arrayrulewidth}
            \end{tabular}
        }
    \caption{Effect of MLLM variants on QVHighlights \textit{val} set.}
    \label{table:other mllms}
\end{table}

\noindent\textbf{MLLM Variations.}
Table~\ref{table:other mllms} shows results using two recent MLLMs: LLaVA~\cite{llava2023} and InternVL2~\cite{internvl2024}. InternVL2 performs best overall, while LLaVA remains competitive and exceeds InternVL2 on R1@0.5. These results highlight the robustness of our method and its effectiveness in leveraging scene understanding for query-aware filtering across different MLLMs. \\

\noindent\textbf{Limitations.} We incorporated MLLMs into MR/HD tasks to enhance scene understanding with richer external knowledge. However, this increases inference time and the number of parameters. Our future work will focus on reducing reliance on MLLMs during inference and finding ways to utilize caption knowledge without directly using captions.


\section{Conclusion}
\label{sec:conclusion}
We propose an important word-aware clip filtering framework to imporve MR and HD tasks by focusing on the most relevant information in video content. We adopt MLLM to fully leverage the knowledge of the captions from each video clip to further understand the video. Our approach includes a feature enhancement module to identify/prioritize important words and enhance the semantic understanding of video clips, while a ranking-based filtering module iteratively refines video clips based on their relevance to the query. This results in improved performance on MR and HD tasks.

\section{Acknowledgments}
This work was supported by the NRF grant funded by the Korea government (MSIT) (No. RS-2023-00252391), and by IITP grant funded by the Korea government (MSIT) (No. RS-2022-00155911: Artificial Intelligence Convergence Innovation Human Resources Development (Kyung Hee University), No. RS-2025-25442384, IITP-2023-RS-2023-00266615: Convergence Security Core Talent Training Business Support Program, No. RS-2022-II220124, Development of Artificial Intelligence Technology for Self-Improving Competency-Aware Learning Capabilities).

\bibliography{aaai2026}

\setcounter{section}{0}
\setcounter{figure}{0}
\setcounter{table}{0}
\setcounter{equation}{0}
\pagenumbering{gobble}

\maketitlesupplementary

\noindent In this supplementary material, we provide additional details and further validate the effectiveness of the proposed method by providing as follows:
\begin{itemize}
    \item \textbf{Weighted Summation Process of Caption.}
    \item \textbf{Additional Details about Caption.}
    \item \textbf{Effect of Similarity Fusion.}
    \item \textbf{Computational Costs.}
    \item \textbf{Results on the Youtube HL.}
    \item \textbf{Additional Visualization.}
\end{itemize}

\section{Weighted Summation Process of Caption}

In this section, we provide a detailed description of the weighted summation process of caption and text query before the caption is input to the feature enhancement module (FEM) after passing the encoder, as described in Section ``Proposed Method'' of this main paper. Note that, the FEM takes caption features, visual features, and query features as input.
The original caption features $F_{oc} \in \mathbb{R}^{L_v \times L_c \times d}$ that pass through the encoder and projection layer are aligned with the query features $F_q \in \mathbb{R}^{L_q \times d}$ using the weighted summation mechanism. This method computes attention weights between the two modalities, leveraging their similarity to refine the original caption features. The caption features $F_c \in \mathbb{R}^{L_v \times d}$ output through this process effectively encapsulate the information of the text query.

First, we expand the query features to match the dimension of the original caption features, generating $F_q^\text{expanded} \in \mathbb{R}^{1 \times L_q \times d}$. Then, we compute the similarity between the original caption features of each clip and the query features.
\begin{gather}
    \text{Sim}(F_{oc}, F_q) = F_{oc} \cdot (F_q^{\text{expanded}})^\top \in \mathbb{R}^{L_v \times L_c \times L_q}.
\end{gather}

We aggregate the similarity across the text query dimension $L_q$ to calculate attention weights $A \in \mathbb{R}^{L_v \times L_c}$ for each caption sequence within a clip. The softmax function ensures that $A$ are normalized across the caption sequence dimension $L_c$.
\begin{gather}
    A = \text{softmax} \left(\frac{1}{L_q} \sum_{L_q} \left( \text{Sim}(F_{oc}, F_q) \right) \right).
\end{gather}

Then, $A$ is reconstructed to fit the $F_{oc}$ dimension and used to perform a weighted sum over the caption sequence of each clip. As a result, the caption features $F_c \in \mathbb{R}^{L_v \times d}$ are generated, which are aligned with the text query.
\begin{gather}
    F_c = \sum_{L_c} A \odot F_{oc},
\end{gather}
where $\odot$ denotes element-wise multiplication, and the summation is performed over the caption sequence dimension $L_c$.

This weighted sum mechanism ensures that the caption features are dynamically refined based on the semantic relevance of the text query, enhancing the model's ability to locate and highlight text-query-related content in video clips.

\section{Additional Details about Caption}

\subsection{Details of Caption Extraction}
\label{B1}
To extract captions for video clips, we employed InternVL2-1B \cite{internvl2024}, a state-of-the-art multimodal large language models (MLLMs) capable of generating concise and semantically rich descriptions. We generate the captions using the following prompt: ``\textit{Please describe the image in one sentence.}''

We design the above-mentioned prompt to generate concise yet detailed descriptions for each clip. By limiting descriptions to a single sentence, we aim to guide the generated content to focus on the key visual elements of each video clip, avoiding unnecessary details while maintaining sufficient context for downstream tasks. As a result, the proposed method effectively leverages not only the visual information from the video but also the textual information, achieving robust performance in moment retrieval and highlight detection.
\begin{figure*}[t]
    \begin{minipage}[b]{0.99\linewidth}
	\centering
        \centerline{\includegraphics[width=\linewidth]{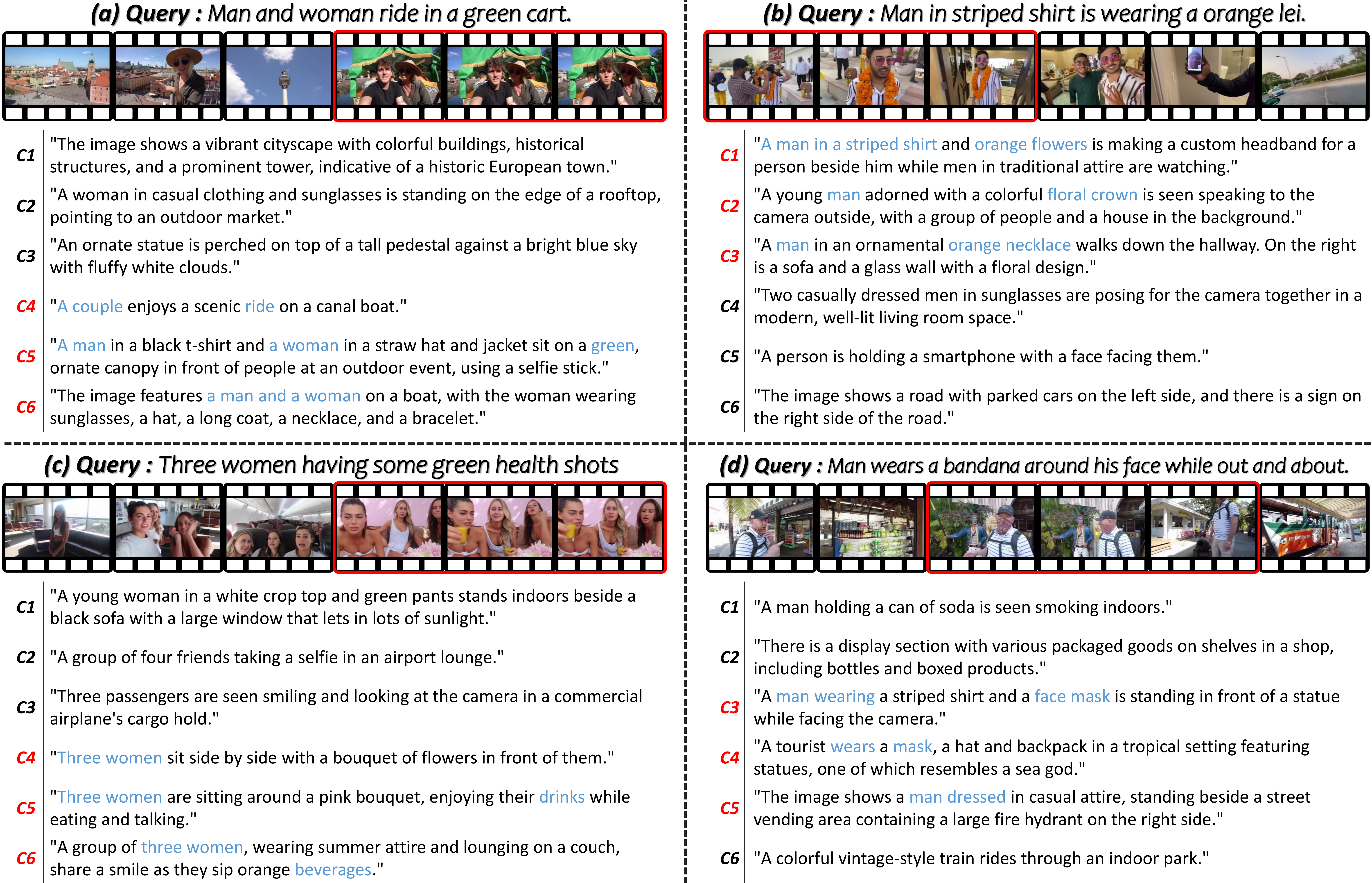}}
        \end{minipage}
    \centering
	\caption{Example captions for the QVHighlights \textit{val} set. C1 through C6 are examples of captions corresponding to six video clips in order. Clips marked with red boxes and text are ground-truth (GT) clips, while the others are non-GT clips.}
    \label{fig:caption visualization}
\end{figure*}
\subsection{Examples of the Generated Caption}
Figure \ref{fig:caption visualization} shows examples of captions generated for video clips on QVHighlights \textit{val} set using the extraction method. We visualize 6 examples of clips from 4 videos: The three clips marked in red are included in the ground truth (GT), and the other three clips are not included in GT.

Specifically, in the GT clips, the generated captions effectively capture the key elements described in the query. For example, for the query ``\textit{Man and woman ride in a green cart.}'' of (a), the caption describes ``\textit{A man in a black t-shirt and a woman in a straw hat and jacket sit on a green, ornate canopy ...}'', which closely matches the semantic requirements of the query. Also, for the query``\textit{Three woman having some green health shots}'' of (c), the caption ``\textit{Three women are sitting around a pink bouquet, enjoying their drinks while eating and talking.}'' accurately identifies the visual elements of the scene.

Captions generated from non-GT clips often contain irrelevant details that do not match the query requirements. For example, in the query ``\textit{Man in striped shirt is wearing a orange lei.}'' of (b), the non-GT caption describes ``\textit{Two casually dressed men wearing sunglasses posing for the camera ...}'', which is irrelevant to the query.

These examples suggest that providing semantically rich captions enables robust moment retrieval and highlight detection.

\section{Effect of Similarity Fusion}
\begin{table}[t!]
    \renewcommand{\tabcolsep}{1.8mm}
    \centering
	\resizebox{0.99\linewidth}{!}{
		\begin{tabular}{c c c ccccc}
            \Xhline{3\arrayrulewidth}
            \rule{0pt}{10.0pt}
            \multirow{3}{*}[-0.8ex]{$\bf{F_{ev}}$} & \multirow{3}{*}[-0.8ex]{$\bf{F_{ec}}$} & \multirow{3}{*}[-0.8ex]{\bf Type} & \multicolumn{3}{c}{\textbf{MR}} & \multicolumn{2}{c}{\textbf{HD} } \\
            \cmidrule(lr){4-6} \cmidrule(l){7-8}
            & & & \multicolumn{2}{c}{R1} & \multirow{2}{*}[-0.6ex]{\makecell{mAP\\Avg.}} & \multirow{2}{*}[-0.6ex]{mAP} & \multirow{2}{*}[-0.6ex]{HIT@1} \\\cmidrule(l){4-5}
            & & & @0.5 & @0.7 & & \\
            \midrule
            - & - & - & 69.16 & 53.81 & 46.81 & 42.86 & 69.61 \\\cdashline{1-8}
            \cmark & - & - & 69.23 & 54.84 & 46.99 & 43.16 & 70.45 \\ 
            - & \cmark & - & 69.35 & 54.52 & 47.06 & 42.99 & 69.87 \\\cdashline{1-8}
            \cmark & \cmark & A & 69.42 & 54.65 & 47.26 & 43.10 & 71.03 \\
            \cellcolor{gray!20}\cmark & \cellcolor{gray!20}\cmark & \cellcolor{gray!20}F & \cellcolor{gray!20}\bf 70.00 & \cellcolor{gray!20}\bf 55.68 & \cellcolor{gray!20}\bf 48.14 & \cellcolor{gray!20}\bf 43.24 & \cellcolor{gray!20}\bf 71.23 \\
            \Xhline{3\arrayrulewidth}
            \end{tabular}
        }
    \caption{Effect of similarity fusion on QVHighlights \textit{val} set. `A' refers the case where $S_{qv}$ and $S_{qc}$ are simply averaged, `F' denotes our similarity fusion procedure.}
    \label{table:similarity fusion}
\end{table}
As presented in the second paragraph of Section ``Ranking-based Filtering Module'' and in Figure 3 of main paper, we performed the following similarity fusion on the query-video similarity matrix $S_{qv}$ and the query-caption similarity matrix $S_{qc}$.
\begin{equation}
\label{eq: weighted mixing}
    S_{qvc} = W S_{qv} + (1-W) S_{qc},
\end{equation}
where $W$ is a learnable weight matrix that controls the balance between $S_{qv}$ and $S_{qc}$. It aims to dynamically adjust the relative importance between the video clip and the caption depending on the situation.

To verify the effectiveness of similarity fusion that integrates both $F_{ev}$ and $F_{ec}$, we performed an ablation study. As shown in Table \ref{table:similarity fusion}, incorporating only $F_{ev}$ or $F_{ec}$ outperforms the baseline that does not consider the RFM. However, considering both features shows better performance. These results demonstrate that in similarity fusion, $F_{ev}$ provides visual information while $F_{ec}$ complements semantic details, thereby improving cross-modal understanding.

Also, to verify the feasibility of performing similarity fusion rather than simply averaging the two similarity matrices, we performed a comparative experiment about two types: A(Average) and F(Fusion). Note that, `A' simply refers to the averaging $S_{qv}$ and $S_{qc}$, which can be represented as follows:
\begin{equation}
    S_{qvc} = \frac{(S_{qv} + S_{qc})}{2}.    
\end{equation}

The results demonstrate that the similarity fusion procedure outperforms simple averaging across all metrics. This indicates that dynamically balancing the importance of video clips and captions allows our model to better adapt to varying contexts, thereby improving performance in both moment retrieval and highlight detection tasks.

\section{Computational Costs}
Table \ref{table:computational cost} compares the training and inference time and the number of parameters of our method with three SOTAs, TR-DETR \cite{trdetr2024}, UVCOM \cite{uvcom2024} and Keyword-DETR \cite{keyworddetr2025}. While caption extraction from the MLLMs adds some time, our method is faster in terms of training time than UVCOM and Keyword-DETR and uses fewer parameters than UVCOM. Additionally, it maintains an efficient inference time of 0.013s (76.9fps), ensuring practicality.
\begin{table}[t!]
    \renewcommand{\tabcolsep}{0.3mm}
    \centering
	\resizebox{\linewidth}{!}{
		\begin{tabular}{l cccc}
            \Xhline{3\arrayrulewidth}
            \rule{0pt}{10.0pt}{\qquad\qquad\textbf{Method}}
            & \makecell{\textbf{MLLMs (s)} \\ \textbf{(\textit{per frame})}}
            & \makecell{\textbf{Train. (s)} \\ \textbf{(\textit{per iter})}} 
            & \makecell{\textbf{Infer. (s)} \\ \textbf{(\textit{per video})}} & \textbf{$\#$ params} \\
            \hline
            \rule{0pt}{9.5pt}TR-DETR \cite{trdetr2024} & N/A & 0.139 & 0.007 & 8.30M \\
            UVCOM \cite{uvcom2024} & N/A & 0.381 & 0.010 & 18.19M \\
            Keyword-DETR \cite{keyworddetr2025} & N/A & 0.804 & 0.027 & 8.32M \\
            \cdashline{1-5}
            \rule{0pt}{9.5pt}\cellcolor{gray!20}\textbf{Proposed Method} & \cellcolor{gray!20}0.770 & \cellcolor{gray!20}0.234 & \cellcolor{gray!20}0.013 & \cellcolor{gray!20}10.61M \\
            \Xhline{3\arrayrulewidth}
            \end{tabular}
        }
    \caption{The comparisons of training time, inference time, and the number of parameters.}
    \label{table:computational cost}
\end{table}


\section{Results on the Youtube HL}
\begin{table}[t!]
    \renewcommand{\tabcolsep}{0.2mm}
    \centering
	\resizebox{\linewidth}{!}{
		\begin{tabular}{lccc cccc}
            \Xhline{3\arrayrulewidth}
            \rule{0pt}{10.0pt}\qquad\qquad\bf Method & \bf Dog & \bf Gym & \bf Par & \bf Ska & \bf Ski & \bf Sur & \bf Avg. \\\hline
            \rule{0pt}{9.5pt}
            UniVTG {\scriptsize(ICCV'23)} \cite{univtg2023} & 71.8 & 76.5 & 73.9 & 73.3 & 73.2 & \underline{82.2} & 75.2 \\
            TR-DETR {\scriptsize(AAAI'24)} \cite{trdetr2024} & 72.9 & 76.2 & 75.3 & 74.4 & 73.5 & \underline{82.2} & 75.8 \\
            UVCOM {\scriptsize(CVPR'24)} \cite{uvcom2024} & \underline{73.8} & \underline{77.1} & \underline{75.7} & \underline{75.3} & \underline{74.0} & \bf 82.7 & \underline{76.4} \\\cdashline{1-8}
            \rule{0pt}{9.5pt}
            \cellcolor{gray!20}\textbf{Proposed Method} & \cellcolor{gray!20}\bf 74.3	& \cellcolor{gray!20}\bf 78.3 & \cellcolor{gray!20}\bf 76.6 &\cellcolor{gray!20}\bf 75.7 & \cellcolor{gray!20}\bf 74.2 & \cellcolor{gray!20}\bf 82.7 &\cellcolor{gray!20}\bf 77.0 \\
            \Xhline{3\arrayrulewidth}
            \end{tabular}
        }
    \caption{Results on highlight detection experiments on the Youtube Highlights. Best/second-best results are marked in \textbf{Bold}/\underline{underlined}.}
    \label{table:youtubeHL}
\end{table}
YouTube Highlights dataset \cite{youtubehl2014} is used for HD task, consisting of videos in 6 categories, with 433 videos. The categories are used as queries. Following UniVTG \cite{univtg2023} and UVCOM \cite{uvcom2024}, we use SlowFast+CLIP features. As shown in Table \ref{table:youtubeHL}, our method achieves an overall average (Avg.) improvement of 0.8$\%$ over SOTAs and outperforms them in all categories.

\section{Additional Visualization}
\begin{figure*}[t]
    \begin{minipage}[b]{1.01\linewidth}
	\centering
        \centerline{\includegraphics[width=\linewidth]{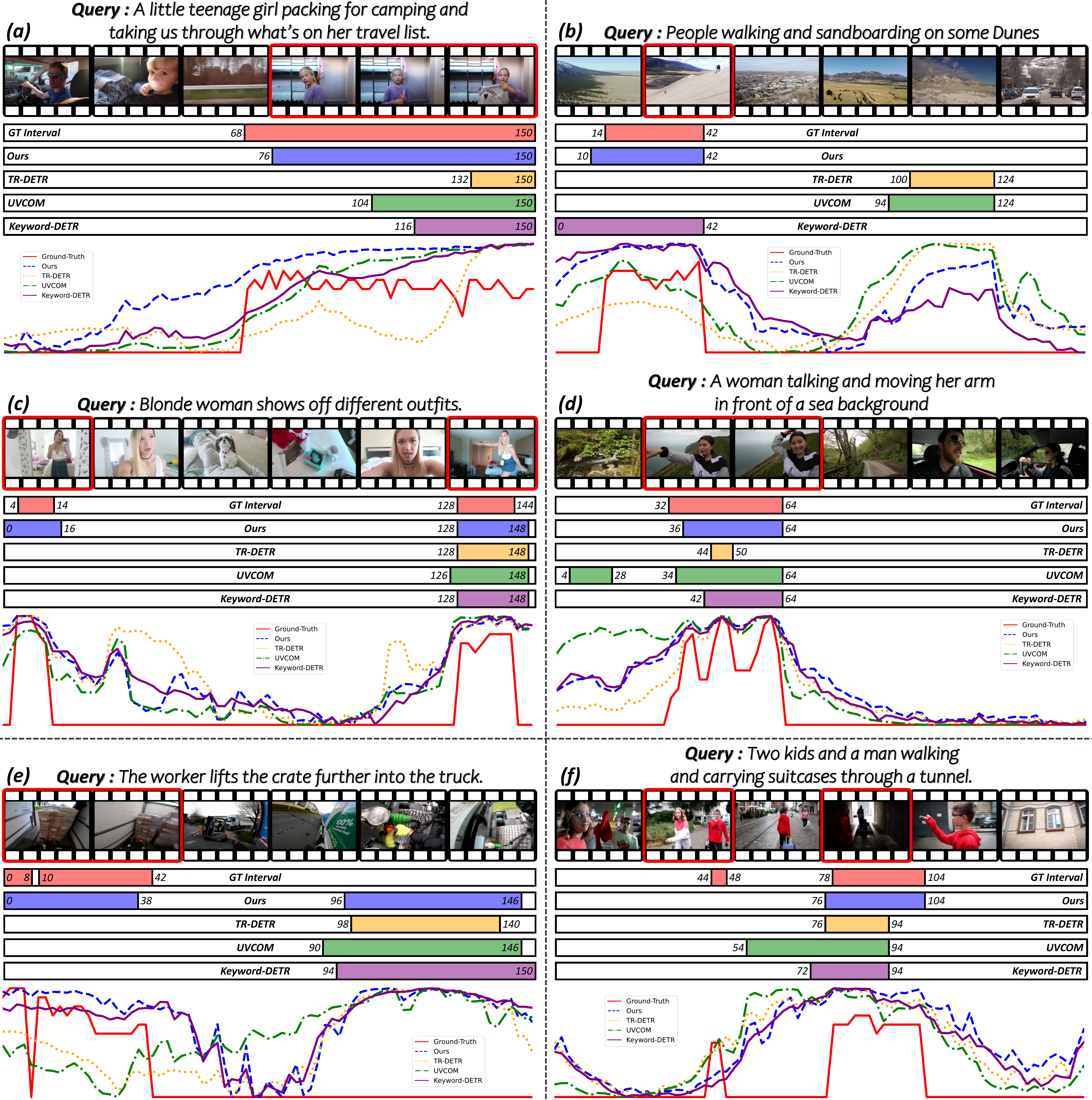}}
        \end{minipage}
    \centering
	\caption{Visualization comparison of MR and HD on QVHighlights \textit{val} set. Prediction results are compared to ground truth (GT), TR-DETR \cite{trdetr2024}, UVCOM \cite{uvcom2024} and Keyword-DETR \cite{keyworddetr2025}. (a) to (d) is examples of correctly predicted results, and (e) to (f) is examples of incorrect prediction results.}
    \label{fig:supple_visualization}
\end{figure*}

Figure \ref{fig:supple_visualization} shows additional qualitative comparisons between the proposed method and TR-DETR \cite{trdetr2024}, UVCOM \cite{uvcom2024} and Keyword-DETR \cite{keyworddetr2025}. The results highlight the superior ability of our approach to distinguish relevant and irrelevant moments and accurately capture the details specified in the query. \\

\noindent\textbf{Success Cases.} In cases where the overall video clip is very similar to the given text query, as in Figure \ref{fig:supple_visualization}(b), our method demonstrates the ability to identify intervals that closely match the ground truth (GT) while avoiding moments that are similar to GT but not GT. On the other hand, other SOTA methods fail to distinguish similar but non-GT clips and incorrectly predict them as relevant moments.

Furthermore, as shown in Figures \ref{fig:supple_visualization}(a), (c), and (d), existing SOTA methods tend to predict only some segments within a GT clip and miss the rest. This is due to structural limitations, such as focusing only on local alignment with the text query or focusing only on segments with strong visual cues. In contrast, the model proposed in this study identifies important words within the text query and iteratively filters relevance across the entire clip based on these important words, enabling more comprehensive predictions across the entire GT clip.\\

\noindent\textbf{Failure Cases.} Figure \ref{fig:supple_visualization}(e) and (f) are examples that were not accurately predicted. In the cases of Figure \ref{fig:supple_visualization}(e), although the GT moment was successfully captured, additional non-GT moment in a similar context were predicted, and in the case of Figure \ref{fig:supple_visualization}(f), most of the GT moment were accurately predicted, but some short additional moments were missed. However, considering that the other methods could not accurately predict any GT moments, this suggests that our method can interpret the meaning of the query more accurately than the other methods.

\section*{References}
\small Chen, Z.; Wu, J.; Wang, W.; Su, W.; Chen, G.; Xing, S.; Zhong, M.; Zhang, Q.; Zhu, X.; Lu, L.; et al. 2024. Internvl: Scaling up vision foundation models and aligning for generic visual-linguistic tasks. In \textit{Proceedings of the IEEE/CVF Conference on Computer Vision and Pattern Recognition}, 24185–24198. \\
\small Lin, K. Q.; Zhang, P.; Chen, J.; Pramanick, S.; Gao, D.; Wang, A. J.; Yan, R.; and Shou, M. Z. 2023. Univtg: Towards unified video-language temporal grounding. In
\textit{Proceedings of the IEEE/CVF International Conference on Computer Vision}, 2794–2804. \\
\small Sun, H.; Zhou, M.; Chen, W.; and Xie, W. 2024. Tr-detr: Task-reciprocal transformer for joint moment retrieval and highlight detection. In \textit{Proceedings of the AAAI Conference on Artificial Intelligence}, volume 38, 4998–5007. \\
\small Sun, M.; Farhadi, A.; and Seitz, S. 2014. Ranking domainspecific highlights by analyzing edited videos. In\textit{ Computer Vision–ECCV 2014: 13th European Conference, Zurich, Switzerland, September 6-12, 2014, Proceedings, Part I 13}, 787–802. Springer. \\
\small Um, S. J.; Kim, D.; Lee, S.; and Kim, J. U. 2025. Watch Video, Catch Keyword: Context-aware Keyword Attention for Moment Retrieval and Highlight Detection. In \textit{Proceedings of the AAAI Conference on Artificial Intelligence}, volume 39, 7473–7481. \\
\small Xiao, Y.; Luo, Z.; Liu, Y.; Ma, Y.; Bian, H.; Ji, Y.; Yang, Y.; and Li, X. 2024. Bridging the gap: A unified video comprehension framework for moment retrieval and highlight detection. In \textit{Proceedings of the IEEE/CVF Conference on Computer Vision and Pattern Recognition}, 18709–18719. \\

\end{document}